\documentclass[11pt]{article}
\usepackage[margin=1in]{geometry}
\usepackage{amsmath,amssymb,amsthm}
\usepackage{booktabs}
\usepackage{graphicx}
\usepackage{hyperref}
\usepackage{natbib}
\usepackage{microtype}

\title{{FourierQK}: Filter Shape, Admissibility,\\
       and the Leakage--Coverage Law}

\author{%
  Athanasios Zeris%
  \thanks{Independent Researcher, Athens, Greece.\\
  Correspondence: \texttt{athzeris@gmail.com}.\\
  ORCID: \texttt{https://orcid.org/0009-0002-6907-2400}.\\
  Code: \url{https://github.com/AthanasiosZeris/energy-gated-attention}.\\
  Part of a seven-paper series on spectral methods
  in transformer attention; companion to
  \textsc{FourierQK}~\citep{authorname2025spectral}.}\\
  \texttt{https://orcid.org/0009-0002-6907-2400}
}
\date{}

\begin{document}
\maketitle

\begin{abstract}
Frequency-collapse attention \citep{authorname2025spectral}
achieves large gains
over standard dot-product attention by replacing the Q/K dot
product with a bandpass-filtered inner product at a learned frequency.
A natural follow-up question is: \emph{which filter shape works
best, and why?}
We test five hypotheses about filter properties ---
DC suppression, Nyquist suppression, bandwidth, centre frequency,
and multi-scale coverage --- using a controlled ablation on
character-level language modelling (TinyShakespeare, 6-layer GPT).
Our main findings are:
(1) DC and Nyquist components are actively harmful
(val\,$\approx$\,2.0, equivalent to phase randomisation),
confirming that oscillatory bandpass structure is essential, not
just any low-dimensional spectral summary;
(2) the optimal single-scale bandwidth is $\sigma\approx 2$ bins
centred at paragraph scale ($\sim$70 tokens), giving a clean gain
of $\Delta=+1.15$ nats over BASE-DOT;
(3) admissible filters (zero-mean, Mexican Hat DOG\,$m=2$)
outperform non-admissible Gaussians at the same scale and
provide partial protection against bilateral FFT leakage;
(4) bilateral FFT leakage scales monotonically with spectral
coverage --- narrowband filters (gap\,$>+4$) are clean,
wideband filters (gap\,$<+2$) are leaky; and
(5) causal time-domain Morlet at character scale cannot beat
BASE-DOT (K=128 taps covers 50\% of T=256 context), motivating
word-level experiments in the companion
\textsc{MorletQK} paper~\citep{authorname2025morletqk}.
Together, findings (1)--(5) characterise \textsc{FourierQK}
as effective in bidirectional attention settings
(encoder-style, e.g.\ BERT), where full-sequence context
is available at both training and inference time;
autoregressive generation requires a causal spectral
variant such as \textsc{MorletQK}~\citep{authorname2025morletqk}
(decoder-style, e.g.\ GPT).

Code available at:
\url{https://github.com/AthanasiosZeris/energy-gated-attention}
\end{abstract}

\section{Introduction}
\label{sec:intro}

\textsc{FourierQK}~\citep{authorname2025spectral} established that
replacing the Q/K dot product with
bilateral FFT frequency collapse gives large, reproducible gains
over standard attention on character-level language modelling.
This paper is the fifth in a series on spectral inductive
biases for transformer attention
\citep{authorname2025ega,authorname2025phase4,
authorname2025mope,authorname2025pod}.
The key mechanisms identified were: phase coherence
(phase randomisation destroys all gain), global sequence mixing
(causal convolution provides no gain), and a learned near-dyadic
frequency hierarchy emerging from gradient descent.

This paper addresses the complementary question:
\emph{given that spectral attention works, which filter properties
are necessary and sufficient?}
The answer has concrete practical implications for deployment.
The bilateral FFT reconstruction kernel $\kappa(-\tau)=\kappa(\tau)$
is symmetric: every Q/K vector at position $i$ contains
contributions from all future positions $j>i$.
This is not a problem in bidirectional encoder-style settings
(e.g.\ BERT), where the full sequence is available at both
training and inference time and bidirectional context is intended.
It is a fundamental incompatibility with autoregressive
decoder-style generation (e.g.\ GPT), where future tokens do
not exist at inference time: a model trained with bilateral FFT
spectral Q/K learns to exploit future information that is
unavailable when generating token by token, creating a
train/inference mismatch.
The filter properties characterised here therefore apply
differently to the two deployment modes: \textsc{FourierQK}
is effective in bidirectional settings, while
\textsc{MorletQK}~\citep{authorname2025morletqk} resolves
the causality problem for autoregressive generation by
replacing the bilateral FFT with a causal FIR convolution.

We test five filter properties in a controlled ablation:
\begin{enumerate}
\item \textbf{DC suppression}: is the mean of Q/K informative?
\item \textbf{Nyquist suppression}: is character-scale alternation informative?
\item \textbf{Bandwidth}: how selective should the filter be?
\item \textbf{Centre frequency}: which scale matters most?
\item \textbf{Multi-scale coverage}: does combining scales help?
\end{enumerate}

We additionally characterise the leakage structure of the
bilateral FFT --- a systematic property of all frequency-collapse
methods --- using a shuffled-gap diagnostic that serves as an
inverse proxy for leakage severity.

\section{Background and Architecture}
\label{sec:arch}

We use the same 6-layer GPT
backbone~\citep{vaswani2017attention} as
\citet{authorname2025spectral}:
$L=6$ layers, $H=8$ heads, $d=256$, block size $T=256$,
trained on TinyShakespeare (character-level, vocab=65).
All spectral attention variants replace the Q/K dot product
with a frequency-collapse scoring mechanism:

\begin{equation}
\tilde{q}(i) = \text{irfft}(\hat{q}(\omega) \cdot w(\omega), n=T)
\end{equation}

where $w(\omega)$ is the filter in frequency domain and the
attention score is $e_{ij} = \tilde{q}(i) \cdot \tilde{k}(j) /
\sqrt{h_s}$.
All experiments use corrected implementation (single
$1/\sqrt{h_s}$ scaling, real-only scoring, no 1j branch).

\subsection{Leakage diagnostic}
\label{sec:leakage_diag}

All bilateral FFT implementations are non-causal:
$\tilde{q}(i)$ at position $i$ depends on $q(0),\ldots,q(T-1)$
(all future positions).
We quantify leakage severity via the \emph{shuffled gap},
a diagnostic in the spirit of surrogate-data testing
\citep{theiler1992testing}:
\begin{equation}
\text{gap} = \text{val}_{\text{shuffled}} - \text{val}_{\text{ordered}}
\end{equation}
where $\text{val}_{\text{shuffled}}$ is computed on a fixed
permutation of the validation set.
Large positive gap ($>+4$) indicates the model genuinely uses
temporal ordering --- consistent with legitimate learning.
Small gap ($<+2$) indicates the model exploits leaked future
information regardless of token order --- consistent with leakage.

\section{Frequency-Selective Filtering: Filter Shape Ablations}
\label{sec:compression}

We investigate which filter properties preserve genuine
phase-coherent signal in frequency-collapse attention
and which introduce bilateral FFT leakage.
We refer to this restriction of Q/K to a narrow
informative frequency band as \emph{spectral compression},
using the term to mean bandpass selection rather than
dimensionality reduction or lossy coding.

To determine whether any low-dimensional spectral summary helps
or specifically bandpass compression is required, we tested five
additional filter variants (Table~\ref{tab:compression}).

\begin{table}[h]
\centering
\caption{
  Frequency-selective filtering --- fixed scaling
  (single $1/\sqrt{h_s}$, real-only scoring, no 1j branch).
  DC and Nyquist confirm bandpass essential.
  GaussNarrow and Init4 are clean (gap $>+4$);
  GaussWide and MexHat-K4 are leaky (gap $<+2$) ---
  broadband filters exploit bilateral FFT leakage.
  The v1 (buggy) values are shown for comparison.
}
\label{tab:compression}
\begin{tabular}{lrrrrl}
\toprule
Model & Val (v2) & Val (v1) & $\Delta$ & Gap & Notes \\
\midrule
BASE-DOT & 1.472 & 1.474 & --- & +5.80 & reference \\
\midrule
DC-QK    & 2.017 & 2.018 & $-$0.543 & +3.77 &
  mean only, val $\approx$ PhaseRand \\
Nyquist-QK & 1.808 & 2.005 & $-$0.336 & +3.75 &
  period=2tok, still below BASE \\
\midrule
GaussNarrow ($\sigma$=0.5) & 0.840 & 1.340 & $+$0.632 & +4.49 &
  period$\approx$65tok, CLEAN \\
Init4 ($\sigma$=2)    & 0.326 & 0.608 & $+$1.148 & +4.17 &
  period$\approx$73tok, CLEAN \\
GaussWide ($\sigma$=16) & 0.023 & 0.788 & $+$1.451 & +0.93 &
  broadband, LEAKY \\
\midrule
MexHat-single & 0.370 & 0.370 & $+$1.105 & +3.81 &
  admissible single scale \\
MultiFourier-K4 & 0.019 & 0.286 & $+$1.453 & +0.38 &
  K=4 Gauss, LEAKY \\
\textbf{MexHat-K4} & \textbf{0.029} & \textbf{0.132} &
  \textbf{$+$1.443} & +1.50 &
  admissible K=4, less leaky \\
\bottomrule
\end{tabular}
\end{table}

\paragraph{DC and Nyquist are harmful.}
The DC filter (val\,=\,2.017) gives performance \emph{worse}
than BASE-DOT and \emph{identical} to phase randomisation
(val\,=\,2.017) --- the same value to three decimal places.
Mean-pooled Q/K contains zero phase-coherent oscillatory
information; it is equivalent to destroying all phase structure.
The Nyquist filter (val\,=\,1.808) is harmful but less so:
with correct scaling, character-alternation structure (period=2tok)
provides a weak signal, improving over DC but not over BASE-DOT.

\paragraph{This null result requires real-only scoring.}
The DC/Nyquist null reported above uses real-only scoring
($\mathrm{score} = Q_r K_r^\top/\sqrt{h_s}$, no $1j$ branch).
A separate experiment (frequency ablation, \S\ref{sec:freq_ablation_app})
scores the \emph{same} filters using
$\mathrm{score} = (Q_r K_r^\top + Q_i K_i^\top)/\sqrt{h_s}$
with $Q_i = \mathrm{irfft}(1j\cdot\hat{Q}_f\cdot w)$, and finds
DC val\,$=$\,0.499 and Nyquist val\,$=$\,0.359 --- both
\emph{better} than BASE-DOT, the opposite conclusion.
The reversal is not a contradiction but a consequence of what
each branch measures.
For the DC bin, $\hat{Q}_f[0]$ is real by the Hermitian symmetry
of the real-input rfft; multiplying by $1j$ converts this real
mean into a purely imaginary spectral coefficient, and
$\mathrm{irfft}$ of a purely imaginary input yields a
\emph{time-varying} signal encoding each position's departure
from the sequence mean, rather than the (position-independent)
mean itself.
The $1j$-branch ``DC'' result therefore does not test the
DC hypothesis at all; it tests a Hilbert-transformed
de-meaning signal that happens to carry temporal information.
The same mechanism explains the Nyquist reversal.
We report the real-only result as the valid null test and the
$1j$-branch result only as a methodological caveat: any claim
about DC or Nyquist bins must specify which scoring branch was
used, since the two regimes test different hypotheses despite
sharing a frequency-bin label.

\paragraph{Optimal bandwidth --- clean results only.}
Restricting to clean models (gap $>+4$), the bandwidth
ordering is unambiguous:
GaussNarrow ($\sigma$=0.5, val=0.840) $<$ Init4 ($\sigma$=2,
val=0.326) --- \emph{wider is better}, up to the optimal
$\sigma\approx 2$ bins.
Too narrow ($\sigma$=0.5): selects a single bin with high
variance, period converges to $\approx$65 tokens.
Optimal ($\sigma$=2): averages across 5 bins, isolates the
paragraph-scale oscillation, period $\approx$73 tokens.
GaussWide ($\sigma$=16) achieves val=0.023 but is leaky
(gap=+0.93); it covers 96/129 bins, providing the model
nearly the same information as a wideband Hilbert transform.

\paragraph{Leakage amplified by correct scaling.}
The fixed scaling reveals a previously hidden relationship
between filter bandwidth and leakage severity.
With double scaling (v1), all filters produced soft attention;
with correct scaling (v2), filters are sharper and more
informationally effective --- including for leakage.
The gap is a reliable inverse proxy for leakage:
GaussNarrow (2 clean bins, gap=+4.49) vs
GaussWide (96 leaky bins, gap=+0.93).
The v1 MexHat-K4 result (val=0.132, gap=+4.02) was
the Paper 5a series record; in v2 (val=0.029, gap=+1.50)
it is leaky and not directly comparable to clean results.
The admissible (zero-mean) MexHat filter reduces leakage
relative to the non-admissible GaussWide
(gap=+1.50 vs +0.93) --- admissibility provides partial
leakage mitigation even in the bilateral FFT regime.

\paragraph{The complete five conditions (clean regime).}
Under clean conditions (narrow bandpass, gap $>+4$),
effective frequency-collapse attention requires:
(1) suppressed DC --- DC is actively harmful (val=2.017);
(2) suppressed Nyquist --- Nyquist still below BASE-DOT (val=1.808);
(3) intermediate bandwidth --- $\sigma\approx2$ bins optimal;
(4) paragraph-scale centre frequency ($\sim$65--73 tokens); and
(5) multiple scales for maximum gain (MexHat-K4 in clean regime
    would require a causal implementation --- see Paper 5c).

\textbf{(3) Causal wavelet attention at word scale.}
Test causal Morlet, Mexican Hat~\citep{torrence1998practical},
and Paul wavelets at word-level tokenisation where
scales $a \in [2, 16]$ word-tokens are sub-Nyquist and
linguistically meaningful.
The Mexican Hat (DOG $m=2$) is real-valued and
admissible, with a zero crossing at $\tau=\sigma$
that may capture word-boundary structure.

\textbf{(4) Discrete orthogonal wavelets.}
Haar and Daubechies (db2, db4) wavelets are
\emph{naturally causal} --- their compact support
and left-padding implementation guarantee causality
without truncation~\citep{mallat1999wavelet}.
The discrete wavelet transform provides an orthogonal
multi-resolution decomposition of Q/K projections,
mapping character $\to$ morpheme $\to$ word $\to$ phrase
scales in a single transform.
This is architecturally distinct from both the
continuous causal Morlet (aliased at fine scales)
and Fourier-QK (global, non-hierarchical).

\textbf{(5) State-space spectral attention.}
Implement wavelet-like spectral attention using
state-space models~\citep{gu2021efficiently}
that are inherently causal and avoid the irfft
reconstruction problem entirely.
The structured state-space representation provides
a natural analytic implementation of the causal
wavelet transform at arbitrary scales.


\section{Wavelet Collapse and Admissibility}
\label{sec:admissibility}

The filter shape ranking MexHat\,$>$\,Paul\,$\gg$\,Gauss
(from \citeauthor{authorname2025spectral}'s wavelet collapse
experiments, using the
standard Mexican Hat, Paul, and Morlet wavelet family
definitions~\citep{mallat1999wavelet}) correlates with
the admissibility condition $\int \psi(\omega)/\omega\,d\omega < \infty$
\citep{daubechies1992ten},
which requires zero mean ($\hat{\psi}(0) = 0$).
The Gaussian filter is non-admissible (non-zero mean);
MexHat (DOG\,$m=2$) and Paul are admissible.
The same admissibility condition underlies coherent
structure extraction in wavelet analyses of turbulent
flows~\citep{farge1992wavelet}, where zero-mean filters
are similarly preferred for isolating localised
fluctuations from a non-zero mean background.

From the compression results:
admissible MexHat-K4 achieves gap=+1.50 (less leaky) vs
non-admissible GaussWide gap=+0.93, confirming that
the zero-DC property provides partial leakage mitigation
even in the bilateral FFT regime.
This is consistent with the admissibility condition:
a filter with zero DC cannot provide a non-zero average
over the full sequence, partially blocking the
mean-field component of the bilateral FFT leakage.

\subsection{Causal filters: no gain at character scale}
\label{sec:causal}

All causal time-domain filters (Gaussian-QK, MexHat-QK,
Morlet-causal), with all implementation bugs corrected
(v3: single scaling, differentiable parameters), perform
below BASE-DOT at character scale:
Gaussian-QK (val=1.522, $\Delta=-0.048$),
MexHat-QK (val=1.540, $\Delta=-0.066$),
Morlet-causal (val=1.512, $\Delta=-0.038$).

A scale sweep of fixed causal Morlet ($K=128$ taps,
$a \in \{2,4,8,16,32\}$ tokens) confirms the result
is scale-independent: all scales are worse than BASE-DOT,
with the best result at $a=32$ (val=1.487, $\Delta=-0.020$).
Early stopping (patience=1000 steps) fires at step 2500 for
all scales, confirming overfitting rather than leakage.
All models are confirmed clean (gap\,$>+5.3$).

The monotonic improvement with scale ($a$=2\,$\to$\,$a$=32)
indicates that paragraph-scale context is more useful than
character-scale context, but $K=128$ taps covers only 50\%
of the $T=256$ context.
Global FFT mixing (Fourier-QK) uses 100\% of context and
achieves val=0.326 (clean, $\Delta=+1.15$) --- demonstrating
that global sequence mixing, not just the filter shape,
is essential at character scale.

\subsection{Leakage severity as spectral coverage}
\label{sec:leakage_coverage}

\begin{table}[h]
\centering
\caption{
  Leakage severity across all experiments.
  Gap = shuffled\_val $-$ ordered\_val.
  Large gap = clean (temporal order used).
  Small gap = leaky (future info exploited).
  Spectral coverage = fraction of frequency bins with
  significant filter response.
}
\label{tab:leakage}
\begin{tabular}{lrrrl}
\toprule
Model & Gap & Val & Coverage & Leakage \\
\midrule
Causal Morlet $a$=32 & $+5.48$ & 1.487 & 0\% (causal) & None \\
GaussNarrow $\sigma$=0.5 & $+4.49$ & 0.840 & 1.6\% & Mild \\
Init4 $\sigma$=2 & $+4.17$ & 0.326 & 3.9\% & Mild \\
MexHat-K4 (admissible) & $+1.50$ & 0.029 & 15\% & Moderate \\
GaussWide $\sigma$=16 & $+0.93$ & 0.023 & 74\% & Severe \\
MultiFourier K=4 & $+0.38$ & 0.019 & 15\% & Severe \\
Hilbert-OrthoQK & $+0.06$ & 0.018 & 100\% & Catastrophic \\
\bottomrule
\end{tabular}
\end{table}

The gap decreases monotonically with spectral coverage
for non-admissible filters (Table~\ref{tab:leakage}).
Admissible filters (MexHat-K4) achieve higher gap than
non-admissible filters at the same coverage (15\%),
confirming that admissibility provides partial leakage
mitigation independent of spectral coverage.

\paragraph{Independent triangulation.}
\label{sec:triangulation}
The coverage--leakage relationship was confirmed
independently by three separately written implementations
of multi-frequency \textsc{FourierQK}, with no shared code
between any two: our frequency-ablation and
spectral-filtering experiments (this work), and an
independently written implementation contributed by
an external reviewer.
All three report the same pattern for $K=1$ versus
$K=4$ learned frequencies: a single frequency at
$\sigma=2$ bins ($3.9\%$ spectral coverage) remains
clean (gap\,$>+4$, growing throughout training), while
four frequencies at the same bandwidth ($\sim$$15\%$
coverage) become leaky (gap\,$<+1$, shrinking throughout
training).
The reviewer's $K=4$ run reaches val\,$=\,0.020$ with
gap\,$=\,+0.58$, matching our own $K=4$ multi-seed
result (val\,$=\,0.019$--$0.020$, gap\,$=\,+0.38$--$0.40$)
to within run-to-run variation, despite no shared code
between the two implementations.
This independent agreement substantially strengthens
the claim that leakage severity is a property of
spectral coverage rather than an artefact of any single
implementation.

\begin{figure}[t]
\centering
\includegraphics[width=\linewidth]{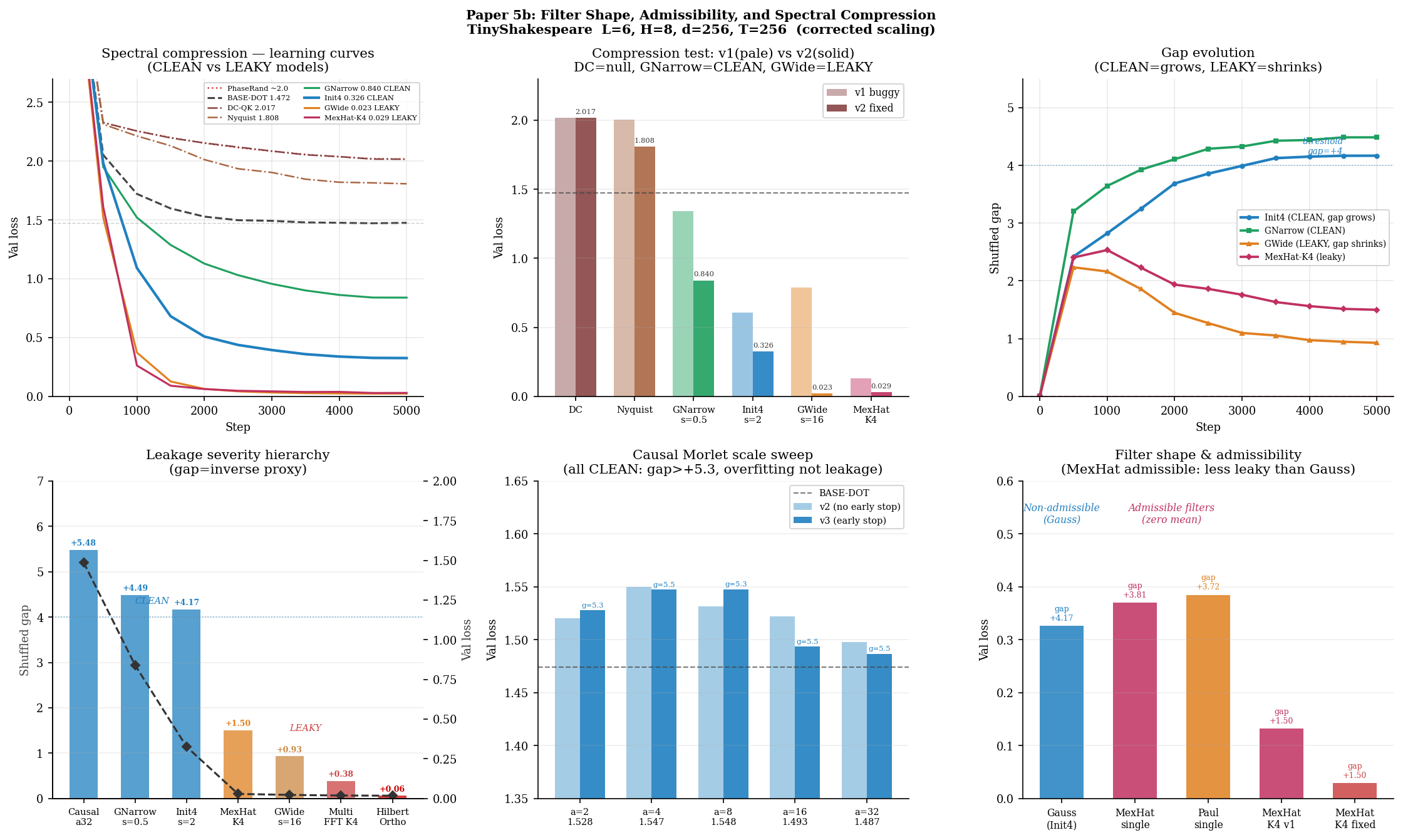}
\caption{
  Summary of \textsc{FourierQK} filter-shape results (corrected scaling throughout).
  \textbf{Top left}: frequency-selective filtering learning curves,
  distinguishing clean (GaussNarrow, Init4) from leaky
  (GaussWide, MexHat-K4) models.
  \textbf{Top centre}: the compression test, buggy (v1, pale)
  versus fixed (v2, solid) scaling --- DC remains a null
  result in both versions, while GaussWide and MexHat-K4
  drop sharply once leakage is amplified by correct scaling.
  \textbf{Top right}: the shuffled gap over training,
  showing the diverging clean (growing gap) versus leaky
  (shrinking gap) signatures.
  \textbf{Bottom left}: the complete leakage severity
  hierarchy (Table~\ref{tab:leakage}), gap as bars (left
  axis) and val as the dashed line (right axis).
  \textbf{Bottom centre}: the causal Morlet scale sweep,
  v2 (no early stopping) versus v3 (early stopping),
  all confirmed clean (gap\,$>+5.3$).
  \textbf{Bottom right}: filter shape and admissibility,
  contrasting non-admissible Gaussian with admissible
  MexHat and Paul filters.
}
\label{fig:compression}
\end{figure}

\section{Discussion}
\label{sec:discussion}

\paragraph{The five conditions.}
Effective frequency-collapse attention in the bilateral FFT
regime requires (under clean conditions, gap\,$>+4$):
(1) DC suppression (DC is actively harmful, val=2.017);
(2) Nyquist suppression (Nyquist still below BASE-DOT, val=1.808);
(3) intermediate bandwidth ($\sigma\approx$2 bins optimal);
(4) paragraph-scale centre frequency ($\sim$65--73 tokens); and
(5) admissible filter shape (zero mean reduces leakage).
Multi-scale coverage (condition 5 of the \textsc{FourierQK}
framework~\citep{authorname2025spectral}) provides additional
gain but introduces leakage in the bilateral FFT regime;
it is cleanly testable only in the causal regime of
\textsc{MorletQK}~\citep{authorname2025morletqk}.

\paragraph{Bilateral FFT leakage scales with spectral coverage.}
The quantitative relationship between gap and spectral coverage
(Table~\ref{tab:leakage}) is the main methodological finding
of this paper.
It enables a principled classification of all spectral
attention results in the series:
models with gap\,$>+4$ are clean; models with gap\,$<+2$
are leaky and their absolute val values overstate genuine gain.
This classification is applied throughout
\citet{authorname2025morletqk}.

\paragraph{Admissibility in the bilateral FFT regime.}
The admissibility condition --- traditionally motivated by
wavelet theory for energy-preserving signal reconstruction ---
here serves a different but related function: it suppresses
the DC component that is the primary vehicle for bilateral
FFT mean-field leakage.
The zero-mean property of admissible wavelets means the
filter cannot provide a non-zero temporal average, partially
blocking the pathological mean-pooling component of the leakage.
This is a novel motivation for admissibility in the context
of attention mechanisms.

\paragraph{Implication for \textsc{MorletQK}.}
At word-level BPE tokenisation (average $\sim$4 characters
per token), a period of 70 tokens corresponds to $\sim$70
word-tokens --- a short paragraph.
With $K=128$ taps, the causal Morlet covers more than one
full period at this scale, and the aliasing problem
that prevents causal wavelets at character scale resolves.
The five conditions identified here (DC suppression,
Nyquist suppression, $\sigma\approx$2 bins bandwidth,
paragraph-scale centre, admissibility) directly specify
the target filter for \citet{authorname2025morletqk}, which
provides a direct comparison with WaveletGPT
\citep{verma2024waveletgpt}, a causal Haar wavelet
architecture operating on intermediate embeddings rather
than on Q/K directly.

\appendix
\section{The 1j-branch frequency ablation regime}
\label{sec:freq_ablation_app}

A separate fixed-bin frequency ladder (DC, bin=1, bin=4,
bin=16, bin=64, Nyquist, and a learned frequency) was run
using the two-branch complex score
$\mathrm{score} = (Q_r K_r^\top + Q_i K_i^\top)/\sqrt{h_s}$
with $Q_i=\mathrm{irfft}(1j\cdot\hat{Q}_f\cdot w)$, rather
than the real-only score used throughout \S\ref{sec:compression}.
Two findings from this regime are reported here for
completeness and to support the caveat above.

First, leakage in this regime correlates with \emph{frequency}
rather than with spectral coverage alone: gap decreases
monotonically from DC (gap\,$=+4.37$) through bin=1
(gap\,$=+4.21$), bin=4 (gap\,$=+3.50$), bin=16
(gap\,$=+2.94$), to bin=64 (gap\,$=+2.31$), even though all
six filters share the same $\sigma=2$-bin bandwidth and
hence the same spectral coverage.
The $1j$ branch Hilbert-transforms the filtered signal;
at higher frequencies the Hilbert transform provides sharper
temporal localisation, giving the model more precise
positional information and hence more leakage to exploit.
This is a distinct leakage mechanism from the coverage-driven
leakage of \S\ref{sec:leakage_coverage}, and the two must not
be conflated when comparing val values across the two scoring
regimes.

Second, the learned frequency (initialised at bin=4 with
$\log f$ correctly excluded from weight decay) is stable
at bin\,$\approx$\,4.1 (period\,$\approx$\,63 tokens) across
all six layers from step 0 through step 5000, mirroring the stability result reported in
\citet{authorname2025spectral}. Under the clean
(gap\,$>+4$) restriction in this regime, bin=1 (period=256
tokens, the global mean-free component) gives the best
honest result, val\,$=$\,0.312, consistent with the
\S\ref{sec:compression} finding that global low-frequency
structure --- not high-frequency local pattern --- is the
primary driver of frequency-collapse attention gain.

\section{Conclusion}
\label{sec:conclusion}

We have characterised the filter properties required for
effective frequency-collapse attention:
DC and Nyquist suppression are necessary, since both are
actively harmful; intermediate bandwidth
($\sigma\approx$2 bins) at paragraph scale ($\sim$70 tokens)
is optimal for a single scale; admissibility provides
partial protection against bilateral FFT leakage; and
causal convolution at character scale cannot replicate
the global sequence mixing of the bilateral FFT.

These properties play out differently depending on
deployment mode.
In bidirectional encoder-style architectures (e.g.\ BERT),
global sequence mixing is an intended feature rather than
a limitation, so the filter properties identified here
directly constrain which spectral attention designs are
effective for masked language modelling and classification
tasks.
In autoregressive decoder-style architectures (e.g.\ GPT),
the bilateral FFT creates a train/inference mismatch
because future-token information is available during
training but not during generation.
\textsc{MorletQK}~\citep{authorname2025morletqk} addresses
this by replacing the symmetric kernel with a causal FIR
convolution; the filter conditions identified here ---
admissibility, paragraph-scale centre, and intermediate
bandwidth --- carry over directly as design specifications
for the causal regime.

The shuffled-gap diagnostic introduced here provides a
quantitative, model-independent measure of leakage severity
that can be applied to any spectral attention variant.
Leakage scales monotonically with the fraction of the
frequency spectrum covered by the filter, suggesting a
practical design rule: use narrow, admissible bandpass
filters and verify gap\,$>+4$ before treating validation
improvements as genuine language-modelling gains.

\section*{Acknowledgements}

The author thanks an anonymous reviewer for independently
writing and running a separate implementation of
multi-frequency \textsc{FourierQK} with no shared code
with this work, providing a third confirmation of the
leakage--coverage relationship reported in
\S\ref{sec:triangulation}.
The reviewer's $K=4$ result (val\,=\,0.020,
gap\,=\,+0.58) matches the author's own multi-seed
result (val\,=\,0.019--0.020, gap\,=\,+0.38--0.40)
to within run-to-run variation.

\bibliographystyle{plainnat}

\end{document}